# Geospatial foundation-model embeddings improve population estimation unevenly across space and scale

Wenbin Zhang[1#], Eimear Cleary[1], Francisco Rowe[2], Somnath Chaudhuri[1], Maksym Bondarenko[1], Shengjie Lai[1#], Andrew J. Tatem[1]

1. WorldPop, School of Geography and Environmental Sciences, University of Southampton, United Kingdom

2. Geographic Data Science Lab, Department of Geography and Planning, School of Environmental Sciences, University of Liverpool, United Kingdom

# Corresponding authors: Wenbin Zhang (wb.zhang@soton.ac.uk), and Shengjie Lai (Shengjie.Lai@soton.ac.uk)

## Abstract

Reliable subnational population estimates are essential for applications, yet remain difficult where censuses are sparse, outdated or spatially coarse. Existing population-mapping workflows rely on hand-built geospatial covariates, such as settlement extent, night-time lights, and environmental conditions, which must be assembled and harmonised across scales and geographies. Geospatial foundation models offer an alternative by learning reusable representations of place from more multifaceted and heterogeneous data sources. Here, we benchmark Population Dynamics Foundation Model (PDFM) embeddings against the harmonised geospatial covariates for subnational population estimation in Brazil, Nigeria and the United States. Under geographically structured validation, PDFM increased predictive fit by a median of 20.1% (IQR: 10.0-33.2%, across country-model comparisons) reduction in unexplained variance, and reduced Kullback-Leibler divergence by 23.2% (9.2-26.2%). However, these gains were uneven. PDFM was most advantageous where the geospatial covariates weakly characterised settlement context, such as larger and less-developed subnational areas. Moreover, PDFM performance was scale-coupled with embeddings providing less flexible transfer across spatial aggregations than geospatial covariates. These findings showed that geospatial foundation-model representations of place can improve population estimation in data poor settings, but their benefits break down predictably under spatial scale mismatch, revealing a fundamental limitation of current geospatial AI.

# Introduction

Accurate small-area population data are increasingly used to inform infrastructure planning, service provision, disaster risk reduction, and public health analysis[1-5]. These estimates may be reported for administrative or statistical units, such as districts, municipalities, counties or census tracts[6,7], or redistributed onto regular grids to create spatially consistent population surfaces[8,9]. In both cases, small-area population products usually depend on census or registry counts, which provide the demographic totals or observations from which finer-scale estimates are derived[10]. However, census and registry data vary widely in quality, spatial detail and update frequency, and can become outdated between enumeration rounds, particularly in rapidly changing settings[11,12]. This creates a persistent need for methods that can use widely available spatial information to refine, update or extrapolate population estimates in settings where reliable small-area counts are unavailable or incomplete.

Population modelling provides one way to address this gap by linking observed population counts to spatial information that is more widely available and more frequently updated. Depending on the application, different models may be used to extrapolate counts to un-surveyed areas, update outdated estimates, or redistribute administrative totals to finer spatial units or regular grids[13-15], associated with hand-engineered geospatial covariates that describe settlement structure, accessibility, land cover, topography, climate and night-time lights[11,16,17]. Although these covariate-based approaches have substantially improved small-area population estimation, they remain limited by the need to manually assemble, harmonise and re-engineer predictor layers across countries, spatial scales and modelling tasks. They also depend on a fixed set of interpretable covariates that may not fully capture the broader spatial context and human-activity signals that distinguish places.

Recent geospatial artificial intelligence (GeoAI) models[18,19] and novel data sources, including mobile-device and social-media activity data, have expanded the signals available for characterising human activity[20-22] and place context[23], moving spatial prediction beyond small sets of predefined, interpretable covariates towards learned, high-dimensional representations of place[24]. Among these, the Population Dynamics Foundation Model (PDFM) provides reusable location embeddings learned from multimodal data that integrate aggregated behavioural and environmental signals across administrative units through a graph neural network[25]. Such representations are appealing for spatial demography because they compress not only otherwise difficult-to-access behavioural signals, such as aggregated search and activity patterns, but also geospatial and contextual data into a single reusable representation of place, thereby reducing the need for downstream applications to assemble and align multiple data modalities separately[26,27]. However, it remains unclear whether, how, and to what extent such learned place representations improve population modelling relative to established geospatial covariates, and in which settings they are most beneficial.

Here, we benchmark PDFM embeddings against 23 standardised geospatial covariates[28] for subnational population estimation in Brazil, Nigeria, and the United States (US). Both population data and two families of predictors are from 2023, and the geospatial covariates are harmonised to the scale of the PDFM embeddings. Using a unified analytical framework, we ask not only whether geospatial foundation-model embeddings improve population estimation, but where, when and at what spatial scale they provide the greatest benefit.

# Results

## Overview of PDFM embeddings and study design

PDFM embeddings and geospatial covariates characterise places in diverse ways. The geospatial covariates used here are explicit predictors derived from satellite and mapped data, which describe the physical, environmental, accessibility and built settlement context, including land cover, topography, climate, roads, night-time lights, and built-up structure[28]. By contrast, PDFM embeddings are learned representations derived from more multifaceted and heterogeneous place-based data streams, including aggregated search trends, maps, busyness (activity levels), weather, and air-quality signals. These inputs are modelled using graph neural networks and aggregated to defined spatial units to produce privacy-preserving, localized embeddings. Thus, PDFM may encode built-environment information, but in a latent form alongside broader digital-footprint and human-activity signals. Figure 1a-f shows the spatial distributions of two PDFM dimensions (Feature 0 under search trend and Feature 151 under maps and busyness) across Brazil, Nigeria, and the US.

Our analysis consisted of three stages as shown in Figure 1g. First, we conducted a benchmark analysis within each country separately, comparing PDFM embeddings and the geospatial covariates as parallel predictor sets for modelling administrative-unit population shares using random forest, XGBoost, and elastic net models, evaluated by coefficient of determination ($R^2$) and Kullback-Leibler (KL) divergence. Second, we assessed geographic transferability by pooling PDFM embeddings and the geospatial covariates across countries and testing how the performance gain of PDFM embeddings over the geospatial covariates varied across held-out higher-level administrative units, i.e., states in Nigeria and Brazil and counties in the US, using leave-one-region-out prediction. Finally, we examined how PDFM embeddings differed from the geospatial covariates and whether the two predictor sets provided complementary information by comparing different combinations of PDFM embeddings and the geospatial covariates, alongside permutation importance analysis to identify the most influential PDFM dimensions.

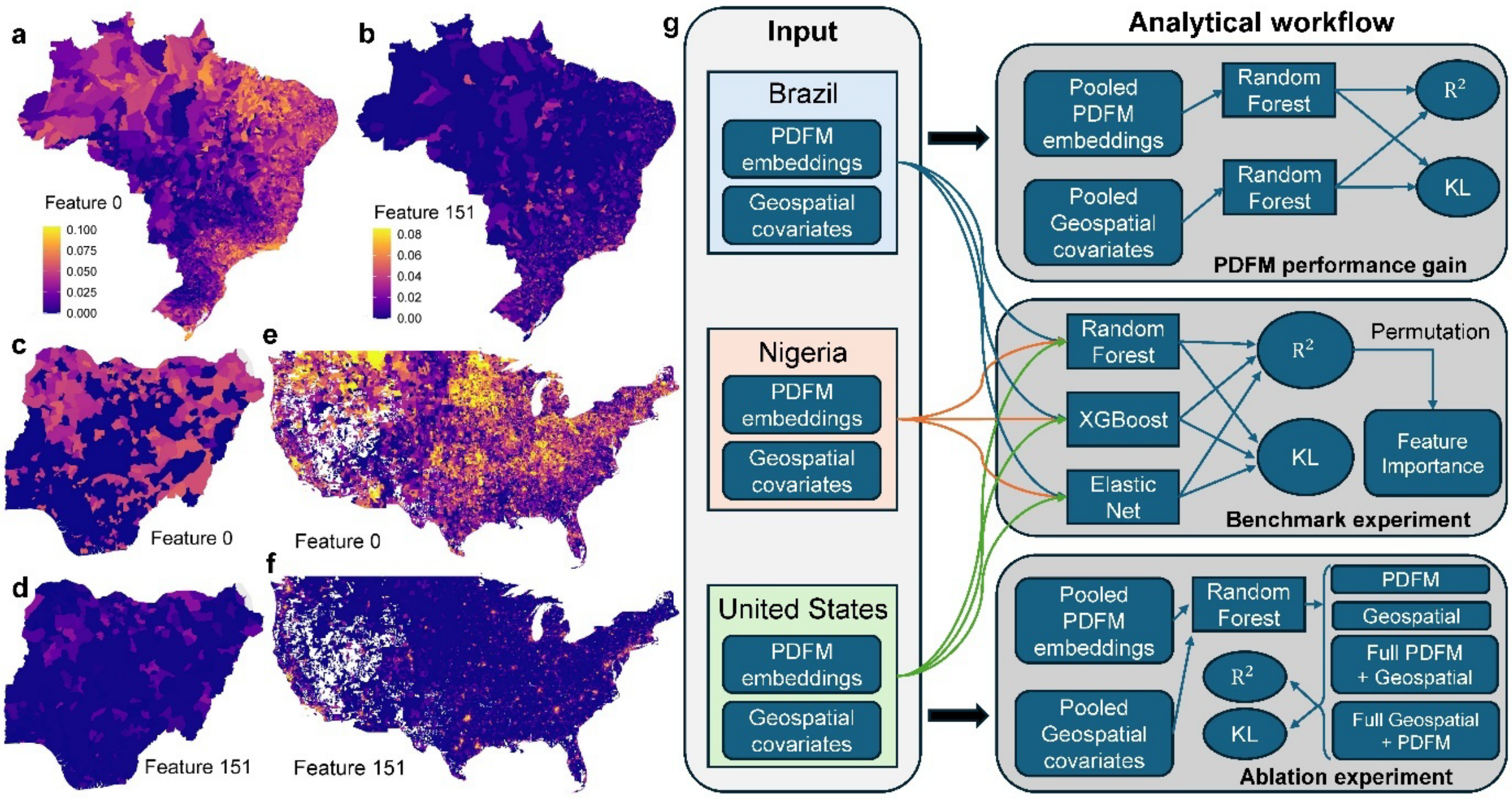

**Figure 1 Overview of PDFM embeddings and study design. a-f,** Example spatial distributions of selected PDFM embedding dimensions across the three study countries. Panels show representative PDFM feature dimensions (Feature 0 and Feature 151) for Brazil **(a,b)**, Nigeria **(c,d)**, and the United States **(e,f)**. **g**, Analytical framework of this study. For each country, PDFM embeddings and the geospatial covariates were used as parallel predictor sets for population modelling at the administrative-unit level. Random forest, XGBoost, and elastic net models were used in benchmark experiment separately within each country. Permutation importance analysis was then used to identify the most influential PDFM dimensions and the geospatial covariates contributing to predictive performance. Additional pooled transferability (PDFM performance gain) and ablation experiments were conducted across countries to evaluate the cross-regional generalisability of PDFM embeddings and the complementary contribution between PDFM embeddings and the geospatial covariates.

## PDFM outperforms geospatial covariates across regions and models

Figure 2a showed that PDFM delivered stronger predictive fit in all country-model combinations, although the size of the advantage varied by geography and model choice. In Brazil, PDFM improved the coefficient of determination ($R^2$) for random forest (median 0.430 [IQR: 0.375-0.490] versus 0.287 [0.247-0.392] for the geospatial covariates), XGBoost (0.406 [0.322-0.503] vs -0.494 [-0.957-0.223]), and elastic net (0.372 [0.318-0.460] vs 0.271 [0.227-0.384]). The same pattern held in the US, where PDFM outperformed the geospatial covariates for random forest (0.758 [0.751-0.766] versus 0.673 [0.662-0.680]), XGBoost (0.607 [0.597-0.615] versus 0.412 [0.405-0.419]), and elastic net (0.726 [0.719-0.731] versus 0.370 [0.353-0.386]). Nigeria also showed consistent gains from PDFM across all three

model classes, including random forest (0.188 [0.138-0.237] versus 0.132 [0.065-0.216]), XGBoost (0.163 [0.112-0.244] versus 0.101 [0.029-0.202]), and elastic net (0.174 [0.072-0.235] versus 0.082 [0.047-0.199]). However, these improvements were smaller in magnitude than those observed in Brazil and the United States.

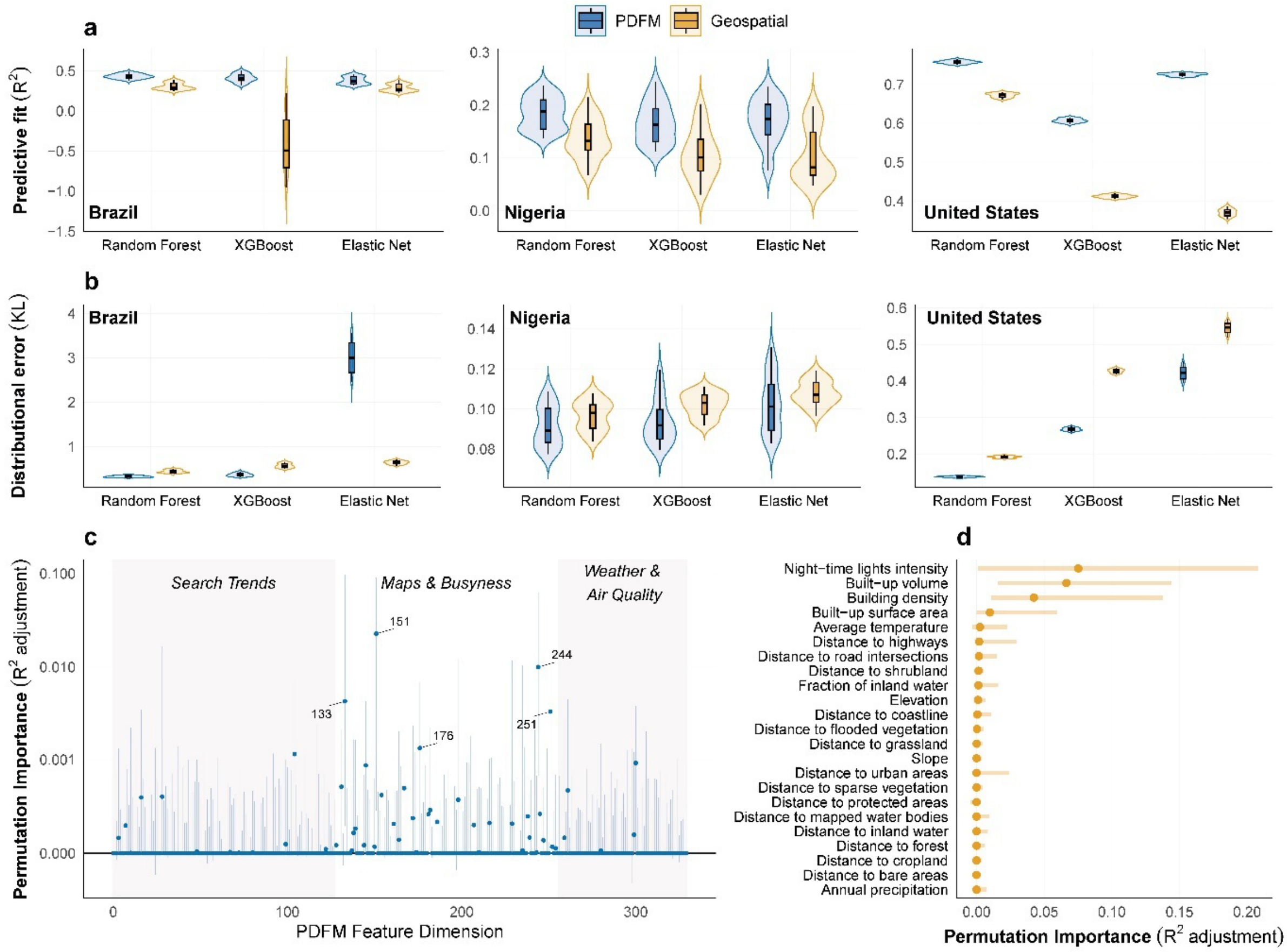

**Figure 2 Benchmarking PDFM features in population extrapolation. a**. Predictive fit ($R^2$) and **b**. distributional error (Kullback-Leibler (KL) divergence on population shares) for models using PDFM embeddings or the geospatial covariates across Brazil, Nigeria, and the US, evaluated with random forest, XGBoost, and elastic net. Violin plots show the distribution across bootstrap replicates; embedded boxplots indicate the median and interquartile range. Higher $R^2$ and lower KL indicate better performance. **c**. Pooled marginal permutation importance of PDFM features across countries and models, expressed as the reduction in $R^2$ after permuting each feature. Shaded regions denote the three semantic feature groups: Search Trends (features 0-127), Maps & Busyness (features 128-255), and Weather & Air Quality (features 256-329). Labels mark the five highest-importance embedding dimensions. **d.** Pooled marginal permutation importance of the harmonised geospatial covariates across countries and models. Because many predictors are correlated and several intervals overlap zero, these estimates should be interpreted as descriptive evidence of broad concentration of predictive signal rather than as a stable rank ordering of individual features.

A similar, though not identical, pattern was observed for the distributional allocation of population shares, measured by KL divergence in Figure 2b, where lower values indicate better agreement. In Brazil, PDFM reduced KL for random forest (0.326 [0.299-0.364] versus 0.435 [0.391-0.501]) and XGBoost (0.361 [0.316-0.442] versus 0.567 [0.486-0.679]), but not for elastic net, where PDFM exhibited a clear failure mode (2.996 [2.444-3.562] versus 0.648 [0.572-0.709]). In Nigeria, differences were modest but consistently favoured PDFM across all models, including elastic net (0.101 [0.083-0.131] versus 0.107 [0.096-0.120]), random forest (0.089 [0.077-0.109] versus 0.098 [0.084-0.108]), and XGBoost (0.092 [0.079-0.120] versus 0.103 [0.092-0.111]). In the US, PDFM consistently achieved lower KL across all three models, including random forest (0.138 [0.134-0.141] versus 0.187 [0.182-0.190]), XGBoost (0.270 [0.264-0.277] versus 0.425 [0.416-0.434]), and elastic net (0.423 [0.394-0.453] versus 0.551 [0.528-0.573]). Taken together, these results indicate that PDFM improved both predictive fit and distributional fidelity, with especially strong and stable gains in the US, broad gains in Brazil, and more modest but consistent improvements in Nigeria.

To explore where predictive signal was concentrated, we pooled marginal permutation importance across countries and models. The results suggested that PDFM importance was unevenly distributed across the 330 embedding dimensions, with a small subset showing higher median reductions in $R^2$ than most other dimensions. However, uncertainty was substantial, and the interquartile ranges for several high-median dimensions included or approached zero; individual feature ranks should therefore be interpreted cautiously. Rather than providing a stable ordering of specific PDFM dimensions, the analysis indicates a sparse pattern of predictive signal, with relatively higher median importance mainly in the Maps & Busyness part and smaller contributions from Search Trends and Weather & Air Quality. The geospatial covariates showed a more interpretable but similarly marginal pattern. Higher median importance was concentrated among settlement-related predictors, especially night-time lights, built-up volume, building density and built-up surface area, whilst most land-cover, terrain and climate variables contributed less.

## Spatial heterogeneity in the performance gains of PDFM

Figure 3 demonstrates that gains from PDFM embeddings were spatially heterogeneous across all settings. For predictive fit ($\Delta R^2$; Figure 3a), Brazil showed consistently positive improvements across most states, whereas Nigeria exhibited a more mixed pattern with both gains and losses. In the US, gains were widespread but spatially patchy, with modest improvements interspersed with localised reversals. Consistent with this spatial heterogeneity, Figure 3b reveals limited evidence for a shared structural gradient underlying $\Delta R^2$ gains, especially in the US. In Nigeria, however, $\Delta R^2$ gains tended to be larger in more extensive subnational areas ($\beta = 0.290$, 95% CI: −0.075 to 0.654), while areas characterised by higher development intensity, captured by higher night-time lights, built-up fraction, building volume, building density, and road accessibility, showed smaller gains ($\beta < 0$ across these indicators, with 95% CIs overlapping or marginally excluding zero). In Brazil, higher PDFM density, reflecting larger average area per embedding, was significantly associated with reduced $\Delta R^2$ gains ($\beta = -2.059$, 95% CI: −3.961 to −0.157), suggesting increasing returns where PDFM signals are dense.

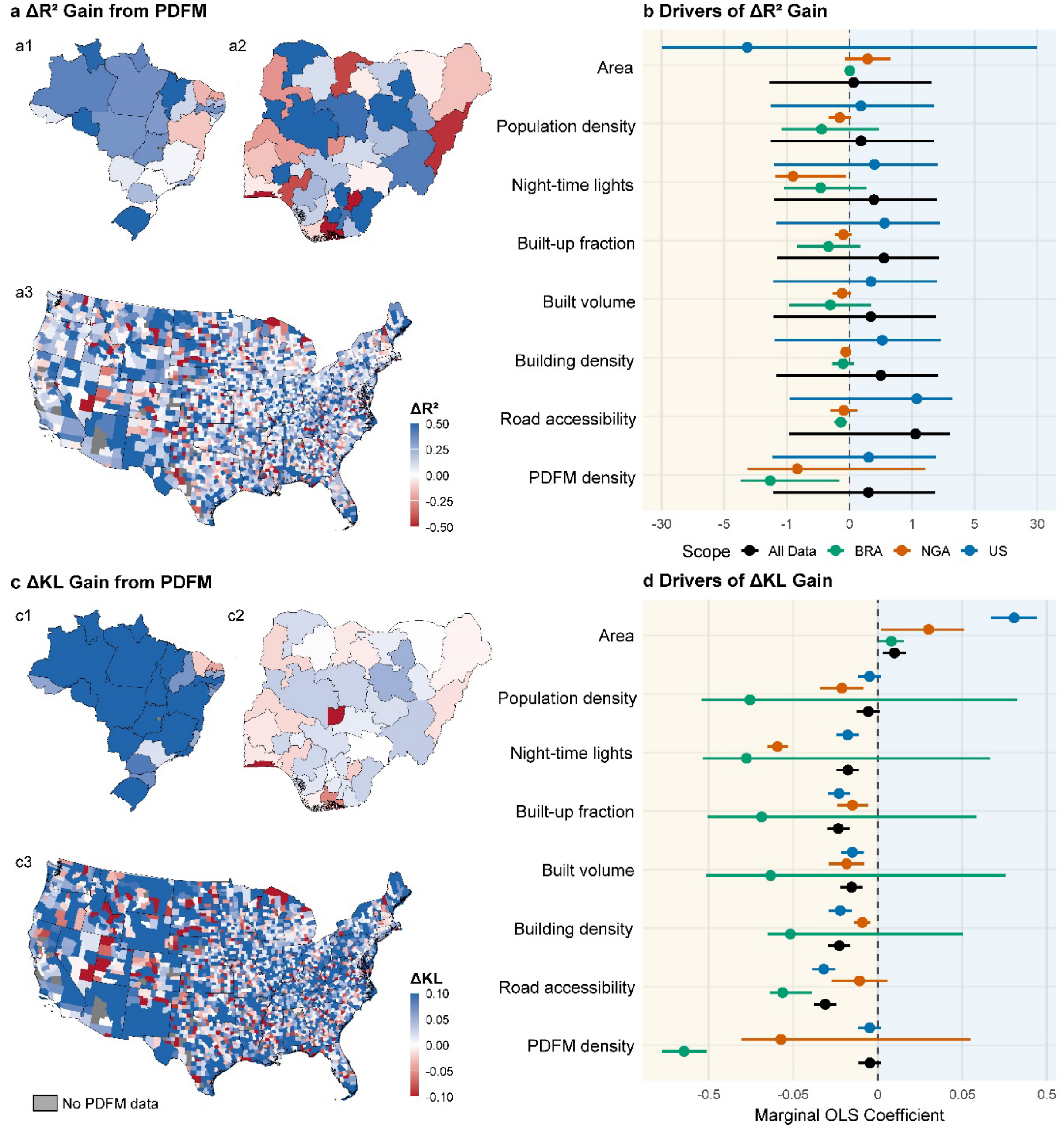

**Figure 3 Spatial heterogeneity and structural correlates of performance gains from PDFM. a**. Spatial distribution of $\Delta R^2$ gain from PDFM in Brazil, Nigeria, and the US, with positive values indicating improved predictive fit relative to models using the geospatial covariates alone. **b**. Marginal ordinary least squares (OLS) coefficients for associations between $\Delta R^2$ gain and candidate structural correlates, shown for the pooled sample and separately for each country. **c.** Spatial distribution of ΔKL gain from PDFM, with positive values indicating improved distributional accuracy. **d.** Marginal OLS coefficients for associations between ΔKL gain and the same structural correlates. Points indicate coefficient estimates and horizontal lines indicate 95% confidence intervals; the dashed vertical line marks zero effect. Negative coefficients indicate that PDFM yields smaller incremental gains in settings where the geospatial covariates might already adequately characterise settlement context, whereas positive coefficients indicate settings where PDFM provides additional

information beyond those standard proxies. Full regression estimates are provided in the Supplementary Information Table 1.

By contrast, PDFM gains in distributional accuracy (ΔKL; Figure 3c) were more systematically structured. Brazil showed near-uniform positive gains from PDFM across states, Nigeria remained regionally heterogeneous, and the US was positive overall despite localised losses. Meanwhile, clearer structural pattern than for $\Delta R^2$ gains is shown in Figure 3d. In general, larger subnational area was associated with greater ΔKL gain (β = 0.010, 95% CI: 0.003 to 0.017), whereas stronger signals in night-time lights (β = −0.018, 95% CI: −0.024 to −0.011), built-up fraction (β = −0.023, 95% CI: −0.030 to −0.017), built volume (β = −0.015, 95% CI: −0.022 to −0.009), building density (β = −0.023, 95% CI: −0.029 to −0.016), and road accessibility (β = −0.031, 95% CI: −0.038 to −0.025) were all associated with smaller gains across countries. With respect to each country, the US showed stronger association regarding area (β = 0.326, 95% CI: 0.203 to 0.449); while Nigeria was more sensitive to population density (β = −0.021, 95% CI: −0.034 to −0.008) and night-time lights (β = −0.133, 95% CI: −0.187 to −0.080). In Brazil, only area, road accessibility and PDFM density are significant, with smaller ΔKL gains were observed in areas with higher road accessibility (β = −0.107, 95% CI: −0.174 to −0.039) and higher PDFM density (β = −1.804, 95% CI: −3.002 to −0.606).

## A small subset of PDFM features captures most predictive signals

Figure 4 demonstrates a clear asymmetry between PDFM features and the geospatial covariates in both predictive fit ($R^2$) and distributional accuracy (KL), highlighting differences in sufficiency and complementarity between the two feature families. Figure 4a and 4b show that adding the geospatial covariates to a model already containing the full PDFM feature set yields negligible gains. Using the full PDFM embeddings only achieved a median $R^2$ of 0.504 (IQR: 0.286-0.568), which was close to the performance of using both the two feature families (median 0.512 [IQR: 0.295-0.581]). Similarly, KL divergence remained stable, with 0.205 (0.188-0.258) for using the full PDFM embeddings compared to 0.197 (0.181-0.248) for using both feature sets. In contrast, using a single geospatial covariate, it achieved a median $R^2$ of 0.405 (0.260-0.465) and KL of 0.419 (0.396-0.452). Performance improved rapidly with additional geospatial covariates, reaching a peak $R^2$ of 0.534 (0.401-0.597) before declining toward 0.475 (0.291-0.574) when all geospatial covariates were included. At the same time, KL decreased markedly to 0.243 (0.228-0.289) but remained consistently higher than the whole geospatial covariates incorporating PDFM features. This non-monotonic pattern suggests that a limited subset of dominant geospatial covariates captures most predictive signal, while additional covariates introduce redundancy and noise.

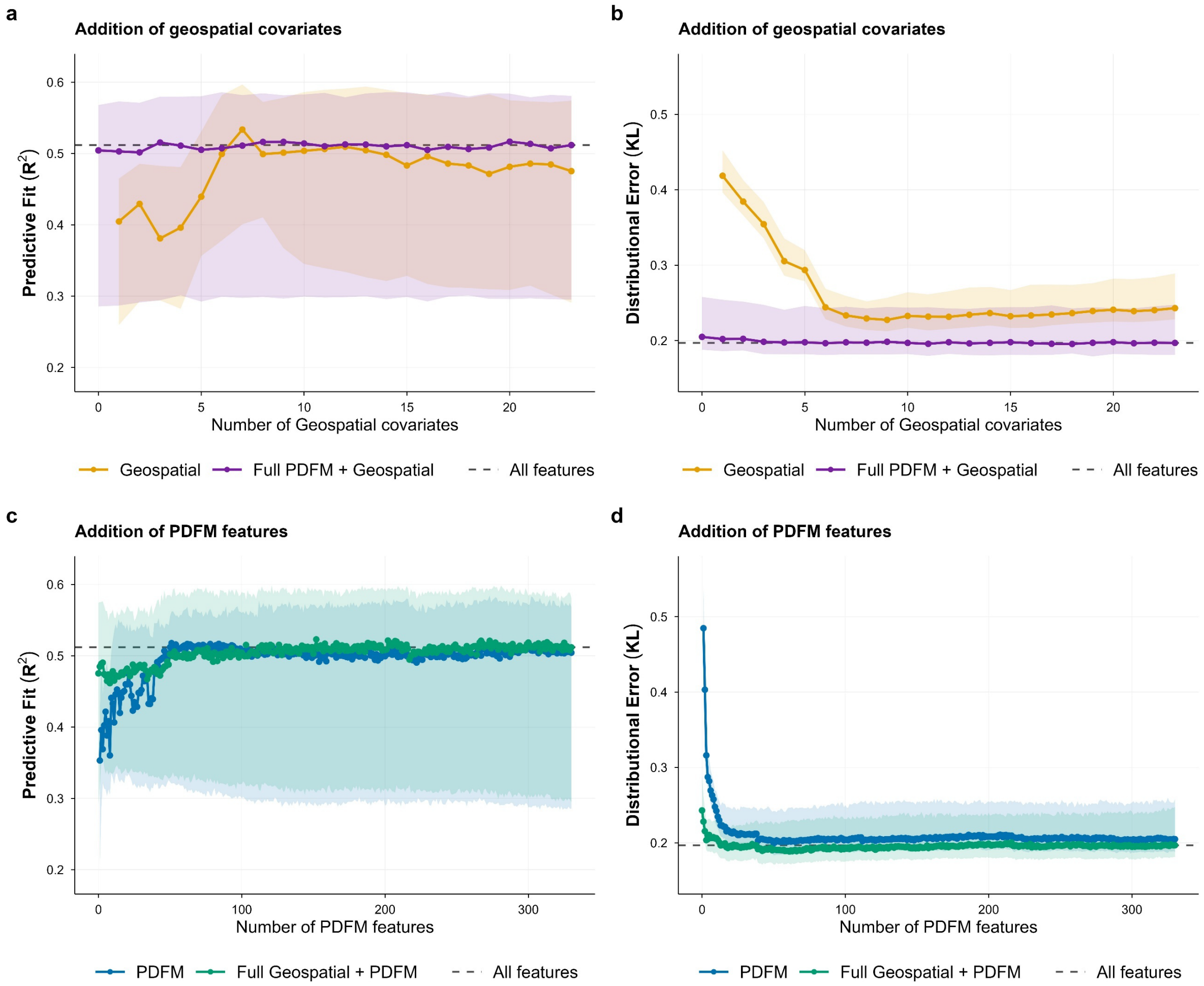

**Figure 4 Marginal contributions of the geospatial covariates and PDFM embeddings to predictive fit and distributional accuracy. a.** Median bootstrap $R^2$ as increasing numbers of the geospatial covariates are included, either alone or in combination with the full PDFM feature set. **b.** Median bootstrap KL divergence under increasing numbers of the geospatial

covariates. **c.** Median bootstrap $R^2$ as increasing numbers of PDFM embeddings are included, either alone or in combination with the full geospatial covariates. **d**. Median bootstrap KL divergence under increasing numbers of PDFM embeddings. Shaded areas show interquartile ranges (from $25^{th}$ to $75^{th}$ percentiles) across 100 bootstrap runs from. The dashed line indicates performance of using all PDFM embeddings and the geospatial covariates.

A more pronounced sufficiency pattern is observed for PDFM embeddings in Figure 4c and 4d. Using a single PDFM embedding had a lower predictive performance and higher distributional accuracy, with $R^2$ of 0.353 (0.207-0.424) and KL of 0.485 (0.458-0.541). However, both metrics improve steadily as additional PDFM embeddings were included. Predictive performance reached a maximum of 0.518 (0.321-0.562) and stabilised near 0.504 (0.286-0.568), while KL declined to 0.205 (0.188-0.258). This indicates that a small subset of PDFM embeddings is sufficient to recover near-maximal predictive accuracy while simultaneously improving distributional fidelity. When PDFM embeddings were added on top of all the geospatial covariates, performance followed a similar but elevated trajectory. It started from a higher baseline of 0.475 (0.291-0.574) in $R^2$ and 0.243 (0.228-0.289) in KL with all the geospatial covariates, and performance improved as additional PDFM features introduced, reaching approximately 0.512 (0.295-0.581) in $R^2$ and 0.197 (0.181-0.248) in KL. Beyond this point, both metrics stabilised, with no evidence of performance degradation from adding lower-ranked features.

## Sensitivity to spatial granularity of modelling units

Figure 5 shows that the predictive advantage of PDFM embeddings was sensitive to the spatial scale. In the main US analysis, conducted at the ZCTA level, PDFM consistently outperformed the geospatial covariates, achieving higher predictive fit and lower distributional error. However, when predictors and population were aggregated to counties and evaluated under states level, this advantage was attenuated and reversed. At the county scale, the geospatial covariates achieved higher median predictive fit than PDFM embeddings, with median $R^2$ of 0.754 (IQR: 0.549-0.876) compared with 0.658 (0.457-0.737) for PDFM.

The geospatial covariates also produced better distributional accuracy, with lower KL divergence than PDFM: 0.113 (0.067-0.181) versus 0.186 (0.149-0.347). This reversal suggests that PDFM embeddings carried their strongest predictive signal at the operational scale at which they were produced, but that this signal did not transfer cleanly after aggregation to coarser administrative units under geographic extrapolation. By contrast, the geospatial covariates appeared more robust in this coarser-scale setting, likely because they can be recalculated or summarised more directly to match the target unit of analysis.

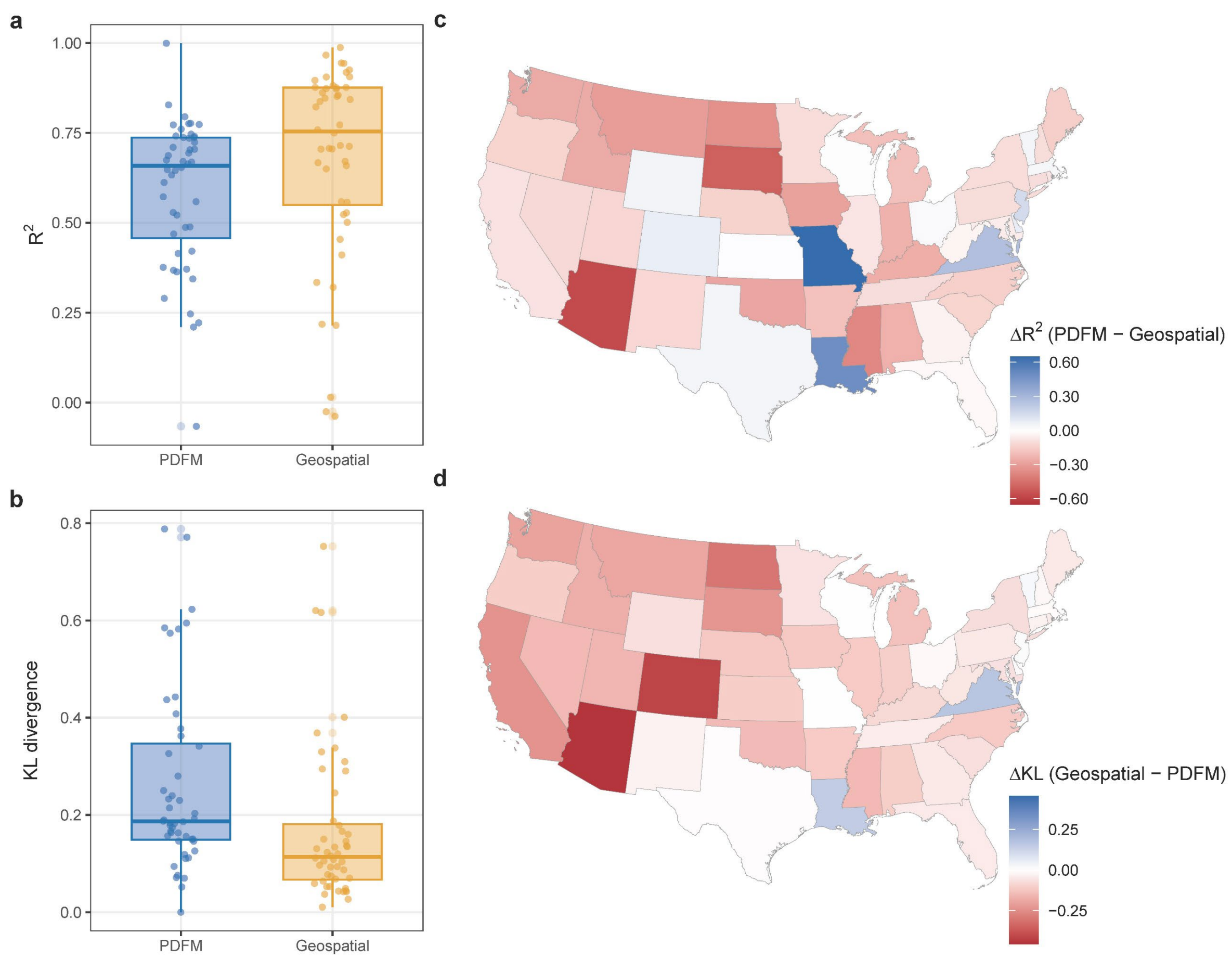

**Figure 5 State-level spatial cross-validation sensitivity analysis in the United States.** Predictors and population were aggregated from ZCTA to county level, and models were evaluated using leave-one-state-out spatial cross-validation, in which all counties within one state were held out jointly and predicted using data from the remaining states. **a**, Distribution of held-out-state $R^2$ values. **b**, Distribution of held-out-state KL divergence values. **c**, State-level difference in predictive fit, where positive values indicate PDFM performs better. **d**, State-level difference in distributional accuracy, where positive values indicate PDFM performs better. Across held-out states, the geospatial covariates showed higher median $R^2$ and lower median KL divergence on average, indicating better representation forms for rescaling information than PDFM embeddings.

## Discussion

This study evaluates the value of PDFM embeddings for spatial population modelling by comparing them with established geospatial covariates across Nigeria, Brazil and the United States. Across country-model comparisons, PDFM improved predictive fit relative to a set of commonly used geospatial covariates by a median of 10.1 percentage points in $R^2$, equivalent to a 20.1% reduction in unexplained variance (IQR, 10.0-33.2%), and reduced KL divergence by a median of 23.2% (IQR, 9.2-26.2%). These gains indicate that PDFM embeddings encode substantial population-relevant spatial information beyond that captured by conventional covariates. However, the advantage was not uniform. The geospatial covariates remained competitive, and in some settings performed better, particularly where explicit built-environment and accessibility predictors already captured settlement patterns well or where predictions were transferred to coarser spatial units. Thus, PDFM embeddings should not be interpreted as universal replacements for established geospatial covariates, but as complementary latent representations whose value depends on geography, spatial scale and transfer setting.

The overall performance gains observed in the national benchmarking analyses are consistent with the design of PDFM embeddings. PDFM is intended to capture complex, multidimensional interactions among human behaviour, environmental conditions and local context at specific locations, providing a rich, location-specific representation of how populations engage with their surroundings[25,29]. These embeddings are broader than the geospatial covariates because they incorporate signals such as regional interests and concerns reflected in search data, activity levels more directly related to the density and frequency of human presence, and geospatial and contextual information about places. Although the geospatial covariates used here are not explicitly replicated by the maps data within PDFM, information related to built-up form, accessibility and settlement intensity may still be represented indirectly through mapped places, activity patterns and contextual signals[30-33]. This is also supported by adding the geospatial covariates to the full PDFM embeddings yielded negligible additional improvement, whereas adding PDFM embeddings to the geospatial covariates improved clear performance. PDFM therefore appears to capture population-relevant information beyond the explicit physical, environmental and settlement covariates used in conventional population-mapping workflows. The gains should consequently be interpreted as evidence of additional representational value, rather than simple superiority over established geospatial covariates, because the two predictor sets characterise places in distinct ways.

Despite the general advantage of PDFM embeddings for population estimation, their superiority was not uniform across subnational regions. Within countries, regions with stronger conventional signatures of development and accessibility tended to show smaller PDFM gains, whereas larger and less-developed subnational regions tended to show greater gains in distributional accuracy. A plausible reason is that, in more developed regions, the geospatial covariates such as built-up intensity, road accessibility and night-time lights has stronger and robust signals already capturing much of the spatial variation relevant to population distribution[34,35], leaving less additional information for PDFM embeddings to contribute. By contrast, where these settlement signals are weaker, sparser, or less complete, PDFM embeddings may add value by capturing broader patterns of human behaviour that are not fully reflected in physical settlement proxies alone[36,37]. The positive association with region size may indicate that, in large administrative units, settlement often occupies only a small fraction of total land area, so the geospatial covariates may be diluted by extensive non-settlement space[16,38], whereas activity-linked human presence signals may remain concentrated in places where people are actually present[25].

Despite multiscale PDFM embeddings can support spatial disaggregation, simple aggregation or disaggregation of the embeddings themselves may be problematic for downstream tasks. PDFM embeddings are a one-off product generated at a particular spatial support encoding specific relational, contextual and activity-linked signals at that support, thus, rescaled embeddings may not preserve all of the information encoded during their production or may alter the dimensional structures among embeddings in ways that reduce their transferability across scales[25,39,40]. By contrast, the geospatial covariates can often be recalculated, averaged, or summed from earth surface to match new target supports under more transparent and intuitive assumptions[10]. This makes the geospatial covariates may retain an important advantage in their relative flexibility across spatial scales, whereas precomputed geospatial foundation-model embeddings may be more tightly coupled to the spatial support at which they are produced.

Several limitations should be acknowledged. First, the study covers three countries and a finite range of administrative supports, so broader evaluation will be needed to determine how robust these patterns are across other geographic and socioeconomic contexts. Second, behavioural and digitally mediated signals are unlikely to be equally representative across places and populations such as children, and this may contribute to uneven model performance over space with diverse demographic structures. Third, the population data are not fully equivalent across countries, and this affects cross-country comparability. Differences in the source and uncertainty structure of the data may influence the apparent relative performance of predictor families. These limitations do not weaken the main contribution of the study, but they do define the conditions under which the findings should be interpreted.

Taken together, these results support a balanced view of geospatial foundation models in spatial population science. PDFM embeddings provide a powerful new representation of population-relevant spatial context and, in our analyses, consistently improved predictive performance relative to the commonly used geospatial covariates. However, their value was conditional rather than universal, depending on whether common indicators of settlement intensity and accessibility were already highly informative. The most promising path forward is therefore not replacement, but evaluation and integration. Hybrid modelling frameworks that smartly combine embedding-based representations with physically grounded settlement indicators and other complementary data sources are likely to provide the most robust basis for accurate and transferable population modelling.

# Methods

## Data collection and processing

**Administrative unit boundary.** Administrative boundaries were defined to align with the spatial resolution of the modelling framework. For Nigeria, we used level-2 administrative units (local government areas, LGAs) from the Global Administrative Areas (GADM) database version 4.1. The administrative hierarchy was constructed directly from GADM attributes, where each level-2 unit (GID_2) is nested within a higher-level unit (GID_1), representing states. For Brazil, we used official district-level boundaries from the Brazilian Institute of Geography and Statistics (IBGE). These correspond to sub-municipal administrative units, and their hierarchical structure (districts nested within states) follows the IBGE regional division framework. For the US, we used ZIP Code Tabulation Areas (ZCTAs) as the primary spatial units. Because some ZCTAs span multiple counties, each ZCTA was assigned to the county containing the largest share of its land area. This results in a two-level hierarchy in which ZCTAs are nested within counties.

**PDFM embeddings.** PDFM embeddings are 330-dimensional feature vectors associated with named locations, produced by the Population Dynamics Foundation Model developed by Google Research, which integrates heterogeneous data sources including aggregated search trends (features 0-127), maps and busyness indicators (features 128-255), and weather and air-quality signals (features 256-329). The embeddings are learned using a self-supervised graph neural network that represents locations (e.g., postal-code and district units) as nodes connected by geographic proximity, administrative relationships and similarity in behavioural signals. During training, the model aggregates multimodal inputs across this graph structure to learn latent representations that capture shared patterns of human activity, infrastructure and environmental context across locations[25]. The resulting embeddings are fixed, location-specific representations that can be used as predictors in downstream tasks. The goal of this study was not to retrain or modify PDFM, but to evaluate the predictive utility of these precomputed embeddings for population modelling relative to established geospatial covariates.

PDFM embeddings were generated separately for Brazil (N = 5568), Nigeria (N = 783) and the US (N = 39649) using the same methodological pipeline. In each country, embeddings were produced from country-specific models built on the relevant administrative geography and available place-based input data. Input signals were time-matched across countries to October 2023 to improve consistency. Consequently, cross-country differences in model performance should be interpreted as reflecting differences in settlement geography, administrative-unit scale, population structure and local input-data coverage, rather than differences in the core embedding methodology. All PDFM records are provided at subnational administrative levels, including local government areas (LGAs) in Nigeria, districts in Brazil, and ZIP Code Tabulation Areas (ZCTAs) in the US, along with their corresponding place names.

However, identical place names can occur in different higher-level regions (e.g., states) in Brazil and Nigeria, which creates ambiguity when linking records to spatial boundaries. To address this, we linked PDFM embedding records to administrative polygons using a combined name-similarity and spatial matching approach. First, all GADM polygons were standardized by converting names to lowercase, removing punctuation, and simplifying common administrative suffixes. For each embedding record, we retrieved geographic coordinates using the Google Maps API based on its Place ID (e.g., ChIJLfXaxFWqNxAR0Do0M-wPyGo). The retrieved PDFM record locations for Brazil are shown in Supplementary Figure 1, and those for Nigeria are shown in Supplementary Figure 2. Candidate polygons were then identified based on spatial proximity to these coordinates, including the nearest polygon and its neighbouring polygons. Within this candidate set, we calculated string similarity between the embedding place name and polygon names using the Jaro-Winkler distance[41]. We evaluated similarity both with the full name and with higher-level region names removed (e.g., removing state names when embedded in district names). The final match was determined by combining spatial proximity and name similarity: the nearest polygon was selected unless another candidate showed substantially higher name similarity beyond a predefined margin. Match quality was assessed by manually reviewing records with low name similarity (10 records in Brazil and 7 records in Nigeria), records whose coordinates or centroids fell outside the matched polygon (4 records in Brazil and 49 records in Nigeria), and records flagged by both criteria (0 record in Brazil and Nigeria).

**Population.** Population data were assembled separately for each country and aligned to the corresponding spatial support of PDFM embeddings. For Nigeria, we used latest population projections for 2022 published by City Population. The population projection assumes the same rate of growth for all LGAs within a state, based on the population of the local government areas (LGAs) of Nigeria according to census results sourced from National Population Commission of Nigeria and National Bureau of Statistics. For Brazil, we used official population data for 2023 published by the Brazilian Institute of Geography and Statistics (IBGE). These data provide authoritative municipality-level population counts, which were aligned to the district-level boundaries used in this study. For the US, we used population estimates from the U.S. Census Bureau's American Community Survey (ACS) 2019-2023 5-year estimates, which are at ZCTAs level derived from survey data.

**Geospatial covariates.** As the comparator predictor family, we used harmonised WorldPop covariates[28] that represent a standardised and widely adopted feature set for spatial population modelling. These covariates consist of raster-based variables capturing key determinants of population distribution, including measures of the built environment, nighttime lights intensity as a proxy for human activity, accessibility indicators, land cover and land use characteristics, as well as environmental and topographic variables such as elevation and slope. All covariates were aggregated to the defined lower-level administrative units to ensure consistency in spatial resolution for modelling. A detailed list of variables and their definitions is provided in the Supplementary Table 2.

## Modelling and evaluation framework

**Models and settings.** All modelling experiments were conducted within a unified framework to enable comparison across predictor sets, countries and experimental settings. Population was modelled at the administrative-unit level as population share, defined as the population in each unit divided by the total population of the study area. To ensure comparability across countries, predictor sets, and resampling iterations, models were fit using fixed settings. Random forest models used 499 trees. XGBoost models used squared-error loss with RMSE as the evaluation metric, 500 boosting rounds, a learning rate of 0.05, maximum tree depth of 6, and row and column subsampling of 0.8. Elastic Net models used a Gaussian error structure with $\alpha = 0.5$.

**Evaluation metrics.** Model performance was assessed using the coefficient of determination ($R^2$) and Kullback-Leibler (KL) divergence. $R^2$ was used to quantify overall predictive fit[42]. KL divergence was used to assess how well the predicted population distribution matched the observed spatial distribution[43]. For KL calculation, a small constant ($\varepsilon = 10^{-12}$) was applied to predicted shares to avoid undefined logarithms. Raw predicted shares were rescaled so that predicted values within each validation set summed to the known total population share of that validation partition. This mass-preserving formulation is consistent with dasymetric and top-down population modelling approaches, in which ancillary predictors are used to estimate relative population weights or densities and are then normalised to known population totals[44].

**Spatial cross-validation.** We used geographically structured spatial cross-validation, where lower-level administrative units were grouped by higher-level administrative regions. In Nigeria, LGAs were grouped by corresponding states; in Brazil, districts were grouped by corresponding states; and in the US, ZCTAs were grouped by corresponding counties. This design reduced information leakage between nearby units and yielded performance estimates that better reflect transferability[45,46].

**Uncertainty.** Uncertainties were estimated by Monte Carlo with 100 iterations. For each analysis, training-validation partitions were defined at the spatial group level. Within each iteration, candidate training sets were generated by sampling approximately 70% of groups. Splits were accepted only when the realised training partition contained 60-80% of lower-level units and 65-75% of total population.

## Country-level benchmarking

The performance of using PDFM embeddings in population modelling was compared with the collected 23 geospatial covariates within each country. Three regression models were evaluated to represent distinct modelling paradigms: Random Forest, XGBoost, and Elastic Net. For each country and resampling split, models were trained separately using either the PDFM embeddings or the geospatial covariates. To characterise model behaviour, we estimated feature importance using permutation importance[47] based on changes in validation $R^2$. Importance scores were calculated for each resampling iteration and model and then aggregated across countries to identify features that were consistently informative for population modelling. Because PDFM dimensions are latent and only partially interpretable, these scores were used to assess the concentration of predictive signal rather than to infer direct mechanisms.

## Place-based transferability analysis

To assess model transferability and identify where different predictor sets perform better, we conducted a pooled cross-region generalisation analysis using a leave-one-out framework. This design tested whether the relative advantage of one predictor family over another persisted under stronger geographic separation than in the within-country resampling analysis. Based on the benchmarking results, only Random Forest was used for this transferability analysis because it showed strong and comparatively stable performance across settings.

Relative transferability was quantified as pairwise differences in $R^2$ and KL between predictor sets for each held-out region, denoted $\Delta R^2 = R^2_{PDFM} - R^2_{Geospatial}$ and $\Delta KL = KL_{Geospatial} - KL_{PDFM}$. Positive values indicate regions where PDFM embeddings outperformed the geospatial covariates, whereas negative values indicate regions where the geospatial covariates remained more competitive. To examine why gains varied across regions, we compiled interpretable regional descriptors including country, total area, population density, night-time lights intensity, built-up fraction, built-up volume, building density, road accessibility, and PDFM density. Associations between these descriptors and $\Delta R^2$ or $\Delta KL$ were examined using univariate ordinary least squares models, i.e., analysing one descriptor at a time, and should be interpreted as exploratory rather than causal.

## Feature sufficiency and complementarity analysis

To determine which predictors were most essential for population modelling, we conducted a pooled feature sufficiency and complementarity analysis across all countries. PDFM embeddings and the geospatial covariates were first ranked separately according to permutation importance from the benchmarking analysis. Predictors were then added progressively from most to least informative to construct a grid of model settings with varying numbers of PDFM embeddings and the geospatial covariates. This yielded three comparison families: PDFM embeddings only, the geospatial covariates only, and combined models containing both predictor types. All three comparison families were evaluated using shared train-validation split per iteration so that performance differences reflected feature composition rather than split variability.

### Sensitivity analyses

To examine whether the performance of PDFM embeddings was sensitive to spatial granularity, we conducted a US county-level sensitivity analysis. Predictors and population were aggregated from ZCTAs to county level, and models were re-evaluated using leave-one-state-out spatial cross-validation. In each fold, all counties within one state were held out jointly and predicted using models trained on counties from the remaining states. The analysis was restricted to the US because aggregation to higher-level administrative units in Brazil and Nigeria would leave too few observations for stable model fitting and spatial cross-validation.

## Data availability

WorldPop geospatial covariate layers are publicly available from the WorldPop geospatial covariate portal: https://hub.worldpop.org/project/categories?id=14. The Google's Population Dynamics Foundation Model (PDFM) embeddings are available for research use (subject to approval) at no cost, more information can be found in https://github.com/google-research/population-dynamics. The Nigeria gridded population data is available from https://www.citypopulation.de/en/nigeria/admin/. For Brazil, district boundary data are available from IBGE's 2022 Census mesh at https://www.ibge.gov.br/en/geosciences/maps/brazil-geographic-networks-

mapasdobrasil/21536-regional-divisions-of-brazil.html, and 2023 municipal population counts are available from IBGE's 2023 municipal population release at https://www.ibge.gov.br/en/statistics/social/population/37757-relacao-da-populacao-dos-municipios-para-publicacao-no-dou-2.html. For the US, ZIP Code Tabulation Area boundaries are available from the U.S. Census Bureau https://www2.census.gov/geo/tiger/TIGER2020/ZCTA520/tl_2020_us_zcta520.zip, the ZCTA-to-county correspondence used for county assignment is available as the 2020 ZCTA-to-county relationship file (www2.census.gov/geo/docs/maps-data/data/rel2020/zcta520/tab20_zcta520_county20_natl.txt), and U.S. population estimates are available from the Census Bureau's 2019-2023 ACS 5-year data (https://www.census.gov/data/developers/data-sets/acs-5year/2023.html).

## Code availability

All analyses were conducted using custom R scripts built around standard open-source packages. The analysis scripts used in this study are available at GitHub: WB-zhang94/Benchmark-PDFM-in-population-modelling.

## Acknowledgements

We thank Google Research for generating and providing access to the experimental time-matched (October 2023) PDFM embeddings, and for continued technical engagement regarding embedding construction and coverage. We acknowledge the WorldPop Program Management Office and Spatial Data Infrastructure team for coordinating data integration and analytical implementation. We also acknowledge the use of the IRIDIS High Performance Computing Facility, and associated support services at the University of Southampton, in the completion of this work. This work was supported by the Patrick J. McGovern Foundation (Grant No. 1556), the Wellcome Trust (Grant No. 308679/Z/23/Z) and the Bill & Melinda Gates Foundation (Grant No. INV-088965). The funders had no role in study design, analysis, interpretation, or manuscript preparation. 

# Supplementary Information

## A. Supplementary Tables

**Supplementary Table 1. Marginal ordinary least squares (OLS) associations between regional characteristics and PDFM performance gains relative to conventional covariates.** Estimated coefficients from marginal ordinary least squares models relating regional characteristics to gains from PDFM over conventional geospatial covariates, analysed separately for predictive fit ($\Delta R^2$) and distributional accuracy ($\Delta KL$) across the pooled sample and within Brazil, Nigeria, and the United States. Positive coefficients indicate that higher values of a given regional characteristic are associated with larger gains from PDFM, whereas negative coefficients indicate smaller gains. Each model includes a single regional characteristic at a time. Reported are the regression estimate, 95% confidence interval, standard error, and two-sided P value. PDFM density reflects the spatial coverage of embeddings, with higher values indicating higher average area represented per embedding.

| Scope | Metric | Variables | Estimate | 95% CI | | Standard error | P value |
|---|---|---|---|---|---|---|---|
| All Data | $\Delta R^2$ | Area | 0.067 | -2.121 | 2.255 | 1.116 | 0.952 |
| | | Population density | 0.185 | -2.003 | 2.373 | 1.116 | 0.868 |
| | | Night-time lights | 0.391 | -1.797 | 2.579 | 1.116 | 0.726 |
| | | Built-up fraction | 0.550 | -1.638 | 2.737 | 1.116 | 0.622 |
| | | Built volume | 0.338 | -1.849 | 2.526 | 1.116 | 0.762 |
| | | Building density | 0.499 | -1.688 | 2.687 | 1.116 | 0.655 |
| | | Road accessibility | 1.225 | -0.963 | 3.412 | 1.116 | 0.272 |
| | | PDFM density | 0.303 | -1.885 | 2.491 | 1.116 | 0.786 |
| | $\Delta KL$ | Area | 0.010 | 0.003 | 0.017 | 0.003 | 0.004 |
| | | Population density | -0.006 | -0.012 | 0.001 | 0.003 | 0.102 |
| | | Night-time lights | -0.018 | -0.024 | -0.011 | 0.003 | 0.000 |
| | | Built-up fraction | -0.023 | -0.030 | -0.017 | 0.003 | 0.000 |
| | | Built volume | -0.015 | -0.022 | -0.009 | 0.003 | 0.000 |
| | | Building density | -0.023 | -0.029 | -0.016 | 0.003 | 0.000 |
| | | Road accessibility | -0.031 | -0.038 | -0.025 | 0.003 | 0.000 |
| | | PDFM density | -0.005 | -0.011 | 0.002 | 0.003 | 0.175 |
| Brazil | $\Delta R^2$ | Area | 0.007 | -0.005 | 0.019 | 0.006 | 0.241 |
| | | Population density | -0.444 | -1.356 | 0.468 | 0.442 | 0.325 |
| | | Night-time lights | -0.460 | -1.188 | 0.268 | 0.353 | 0.205 |
| | | Built-up fraction | -0.333 | -0.838 | 0.173 | 0.245 | 0.187 |
| | | Built volume | -0.306 | -0.962 | 0.351 | 0.318 | 0.346 |
| | | Building density | -0.101 | -0.275 | 0.073 | 0.084 | 0.241 |
| | | Road accessibility | -0.137 | -0.242 | -0.033 | 0.051 | 0.012 |
| | | PDFM density | -2.059 | -3.961 | -0.157 | 0.921 | 0.035 |
| | $\Delta KL$ | Area | 0.008 | 0.001 | 0.016 | 0.004 | 0.036 |
| | | Population density | -0.281 | -0.901 | 0.340 | 0.301 | 0.360 |
| | | Night-time lights | -0.299 | -0.794 | 0.197 | 0.240 | 0.226 |
| | | Built-up fraction | -0.218 | -0.562 | 0.126 | 0.167 | 0.203 |
| | | Built volume | -0.170 | -0.618 | 0.278 | 0.217 | 0.442 |
| | | Building density | -0.066 | -0.184 | 0.052 | 0.057 | 0.262 |
| | | Road accessibility | -0.107 | -0.174 | -0.039 | 0.033 | 0.003 |
| | | PDFM density | -1.804 | -3.002 | -0.606 | 0.580 | 0.005 |
| Nigeria | $\Delta R^2$ | Area | 0.290 | -0.075 | 0.654 | 0.179 | 0.115 |
| | | Population density | -0.156 | -0.336 | 0.023 | 0.088 | 0.086 |
| | | Night-time lights | -0.901 | -1.741 | -0.061 | 0.414 | 0.036 |
| | | Built-up fraction | -0.097 | -0.226 | 0.031 | 0.064 | 0.134 |
| | | Built volume | -0.118 | -0.266 | 0.030 | 0.073 | 0.115 |
| | | Building density | -0.057 | -0.127 | 0.012 | 0.034 | 0.104 |
| | | Road accessibility | -0.090 | -0.303 | 0.122 | 0.105 | 0.395 |
| | | PDFM density | -0.832 | -3.512 | 1.847 | 1.320 | 0.532 |
| | ΔKL | Area | 0.030 | 0.002 | 0.058 | 0.014 | 0.035 |
| | | Population density | -0.021 | -0.034 | -0.008 | 0.006 | 0.002 |
| | | Night-time lights | -0.133 | -0.187 | -0.080 | 0.026 | 0.000 |
| | | Built-up fraction | -0.015 | -0.024 | -0.006 | 0.005 | 0.002 |
| | | Built volume | -0.018 | -0.029 | -0.008 | 0.005 | 0.001 |
| | | Building density | -0.009 | -0.014 | -0.004 | 0.002 | 0.000 |
| | | Road accessibility | -0.011 | -0.027 | 0.006 | 0.008 | 0.196 |
| | | PDFM density | -0.115 | -0.324 | 0.093 | 0.103 | 0.269 |
| US | $\Delta R^2$ | Area | -3.531 | -44.074 | 37.013 | 20.678 | 0.864 |
| | | Population density | 0.183 | -2.040 | 2.405 | 1.133 | 0.872 |
| | | Night-time lights | 0.398 | -1.815 | 2.611 | 1.129 | 0.724 |
| | | Built-up fraction | 0.560 | -1.670 | 2.790 | 1.137 | 0.622 |
| | | Built volume | 0.343 | -1.881 | 2.567 | 1.134 | 0.762 |
| | | Building density | 0.525 | -1.771 | 2.820 | 1.171 | 0.654 |
| | | Road accessibility | 1.302 | -0.953 | 3.557 | 1.150 | 0.258 |
| | | PDFM density | 0.308 | -1.904 | 2.520 | 1.128 | 0.785 |
| | ΔKL | Area | 0.326 | 0.203 | 0.449 | 0.063 | 0.000 |
| | | Population density | -0.005 | -0.012 | 0.002 | 0.003 | 0.162 |
| | | Night-time lights | -0.018 | -0.025 | -0.011 | 0.003 | 0.000 |
| | | Built-up fraction | -0.023 | -0.030 | -0.016 | 0.003 | 0.000 |
| | | Built volume | -0.015 | -0.022 | -0.008 | 0.003 | 0.000 |
| | | Building density | -0.022 | -0.029 | -0.015 | 0.004 | 0.000 |
| | | Road accessibility | -0.032 | -0.039 | -0.025 | 0.003 | 0.000 |
| | | PDFM density | -0.005 | -0.011 | 0.002 | 0.003 | 0.170 |

**Supplementary Table 2. The study geospatial covariates**

| **Covariate** | **Source file** |
|---|---|
| Building density | bra_buildings_count_BCB_gl_100m_v1_1<br>nga_buildings_count_BCB_gl_100m_v1_1<br>usa_buildings_count_PIB_ms_100m_v1_1 |
| Built-up surface area | bra_built_S_GHS_U_wFGW_100m_v1_2023<br>nga_built_S_GHS_U_wFGW_100m_v1_2023<br>USA_Built_surface_2023 |
| Built-up volume | bra_built_V_GHS_U_wFGW_100m_v1_2023<br>nga_built_V_GHS_U_wFGW_100m_v1_2023<br>USA_Built_volume_2023 |
| Distance to coastline | bra_coastline_dst_100m_v1<br>nga_coastline_dst_100m_v1<br>usa_coastline_dst_100m_v1 |
| Distance to inland water | bra_dist_inland_water_100m_esa_2021_v1<br>nga_dist_inland_water_100m_esa_2021_v1<br>usa_dist_inland_water_100m_esa_2021_v1 |
| Elevation | bra_elevation_merit103_100m_v1<br>nga_elevation_merit103_100m_v1<br>usa_elevation_merit103_100m_v1 |
| Distance to cropland | bra_esalc_11_dst_2022_100m_v1<br>nga_esalc_11_dst_2022_100m_v1<br>usa_esalc_11_dst_2022_100m_v1 |
| Distance to forest | bra_esalc_40_dst_2022_100m_v1<br>nga_esalc_40_dst_2022_100m_v1<br>usa_esalc_40_dst_2022_100m_v1 |
| Distance to grassland | bra_esalc_130_dst_2022_100m_v1<br>nga_esalc_130_dst_2022_100m_v1<br>usa_esalc_130_dst_2022_100m_v1 |
| Distance to shrubland | bra_esalc_140_dst_2022_100m_v1<br>nga_esalc_140_dst_2022_100m_v1<br>usa_esalc_140_dst_2022_100m_v1 |
| Distance to sparse vegetation | bra_esalc_150_dst_2022_100m_v1<br>nga_esalc_150_dst_2022_100m_v1<br>usa_esalc_150_dst_2022_100m_v1 |
| Distance to flooded vegetation | bra_esalc_160_dst_2022_100m_v1<br>nga_esalc_160_dst_2022_100m_v1<br>usa_esalc_160_dst_2022_100m_v1 |
| Distance to urban areas | bra_esalc_190_dst_2022_100m_v1<br>nga_esalc_190_dst_2022_100m_v1<br>usa_esalc_190_dst_2022_100m_v1 |
| Distance to bare areas | bra_esalc_200_dst_2022_100m_v1<br>nga_esalc_200_dst_2022_100m_v1<br>usa_esalc_200_dst_2022_100m_v1 |
| Distance to highways | bra_highway_dist_osm_2023_100m_v1<br>nga_highway_dist_osm_2023_100m_v1<br>usa_highway_dist_osm_2023_100m_v1 |
| Fraction of inland water | bra_inland_water_pct_100m_v1<br>nga_inland_water_pct_100m_v1<br>usa_inland_water_pct_100m_v1 |
| Annual precipitation | bra_ppt_2023_yravg_tc_100m_v1<br>nga_ppt_2023_yravg_tc_100m_v1<br>usa_ppt_2023_yravg_tc_100m_v1 |
| Distance to road intersections | bra_rd_intrs_dist_osm_2023_100m_v1<br>nga_rd_intrs_dist_osm_2023_100m_v1<br>usa_rd_intrs_dist_osm_2023_100m_v1 |
| Slope | bra_slope_merit103_100m_v1<br>nga_slope_merit103_100m_v1<br>usa_slope_merit103_100m_v1 |
| Average temperature | bra_tavg_2023_tlst_100m_v1<br>nga_tavg_2023_tlst_100m_v1<br>usa_tavg_2023_tlst_100m_v1 |
| Night-time lights intensity | bra_viirs_fvf_2023_100m_v1<br>nga_viirs_fvf_2023_100m_v1<br>usa_viirs_fvf_2023_100m_v1 |
| Distance to mapped water bodies | bra_waterbodies_dist_osm_2023_100m_v1<br>nga_waterbodies_dist_osm_2023_100m_v1<br>usa_waterbodies_dist_osm_2023_100m_v1 |
| Distance to protected areas | bra_WDPA_pre2022_cat1_dist_100m_v1<br>nga_WDPA_pre2022_cat1_dist_100m_v1<br>usa_WDPA_pre2022_cat1_dist_100m_v1 |

## B. Supplementary Figures

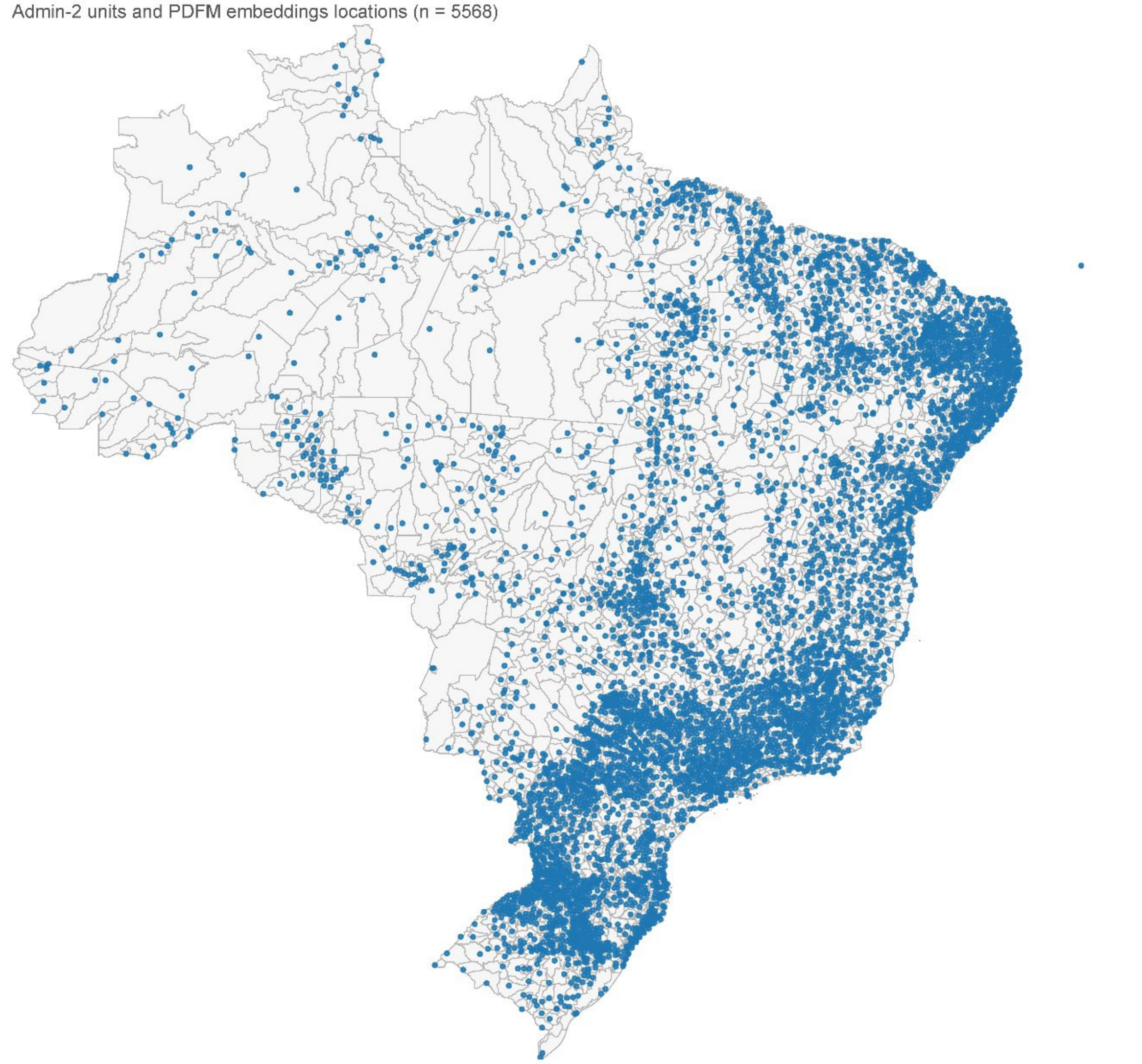

**Supplementary Figure 1. Spatial distribution of PDFM embedding locations in Brazil.** Points indicate the geographic locations of Google Places records used to construct PDFM embeddings (n = 5,568), overlaid on second-level administrative boundaries (municipalities).

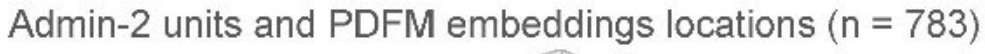

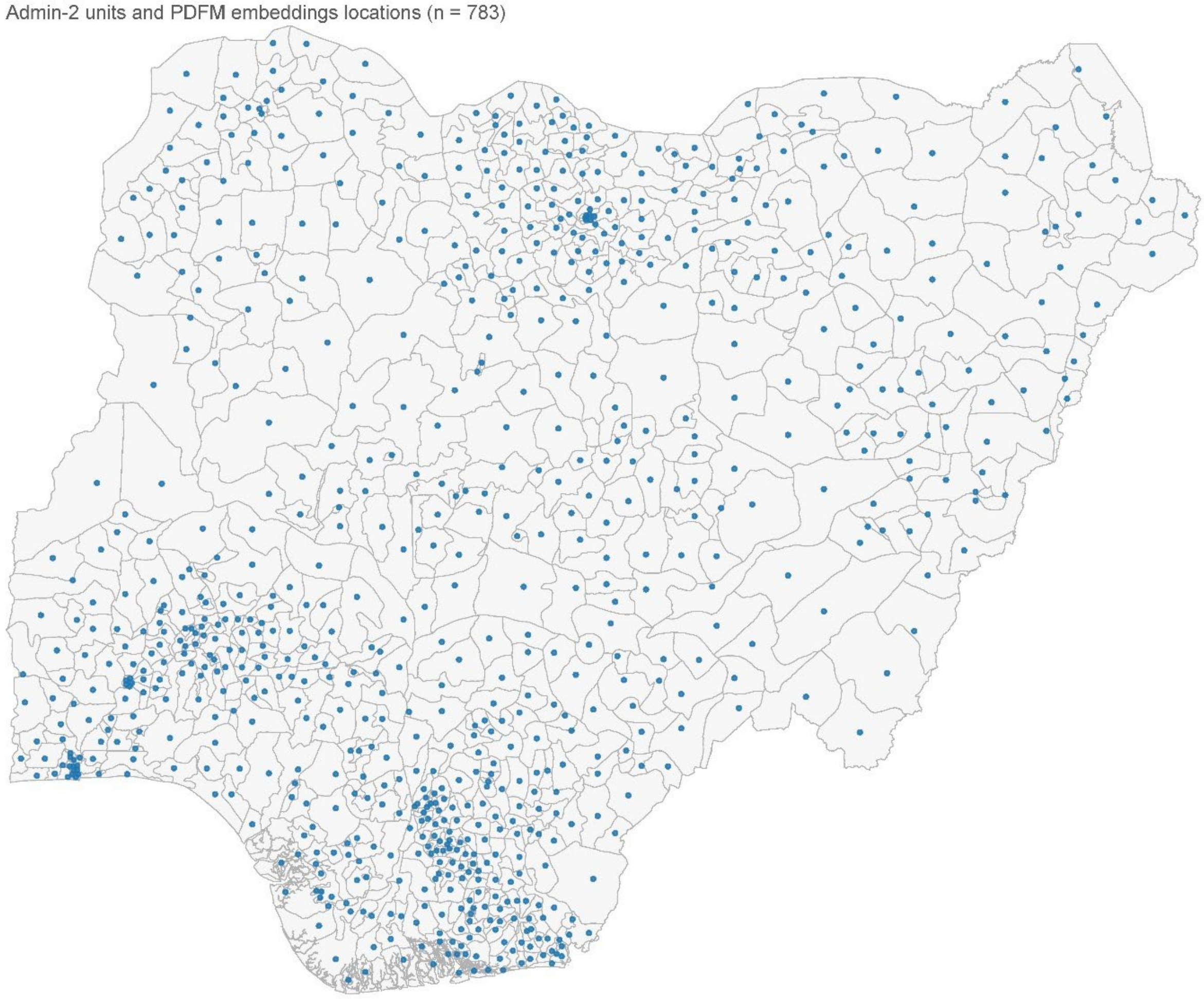

**Supplementary Figure 2. Spatial distribution of PDFM embedding locations in Nigeria.** Points indicate the geographic locations of Google Places records used to construct PDFM embeddings (n = 783), overlaid on second-level administrative boundaries (Local Government Areas).